\documentclass{article} 
\usepackage{iclr2021_conference,times}

\usepackage{amsmath,amsfonts,bm}

\def\eqref#1{equation~\ref{#1}}

\def\1{\bm{1}}

\DeclareMathAlphabet{\mathsfit}{\encodingdefault}{\sfdefault}{m}{sl}
\SetMathAlphabet{\mathsfit}{bold}{\encodingdefault}{\sfdefault}{bx}{n}

\usepackage{hyperref}
\usepackage{url}
\usepackage{hyperref}
\usepackage{amsmath}
\usepackage{geometry} 
\usepackage{enumitem} 
\usepackage{url}
\usepackage{graphicx} 
\usepackage{subcaption} 
\usepackage{floatrow} 
\usepackage{hyperref}
\usepackage{url}
\usepackage{graphicx} 
\usepackage{subcaption} 

\title{A deep learning approach to track eye movements based on events}

\begin{document}
\maketitle

\begin{abstract}
This research project addresses the challenge of accurately tracking eye movements during specific events by leveraging previous research. Given the rapid movements of human eyes, which can reach speeds of 300°/s, precise eye tracking typically requires expensive and high-speed cameras. Our primary objective is to locate the eye center position (x, y) using inputs from an event camera. Eye movement analysis has extensive applications in consumer electronics, especially in VR and AR product development. Therefore, our ultimate goal is to develop an interpretable and cost-effective algorithm using deep learning methods to predict human attention, thereby improving device comfort and enhancing overall user experience.
//
To achieve this goal, we explored various approaches, with the CNN\_LSTM model proving most effective, achieving approximately 81\% accuracy. Additionally, we propose future work focusing on Layer-wise Relevance Propagation (LRP) to further enhance the model's interpretability and predictive performance.

\end{abstract}

\section{Introduction}
Eye-tracking methods have broad applications in domains including consumer electronics, neuroscience and psychology, as it has the potential to understand human visual behavior [1]. Traditional eye-tracking systems typically require specialized hardware to capture eye movements, and analyze massive data in the process. In comparison, event-based eye-tracking only capture sparse data when eye moves, which significantly reduces processing time and power consumption. In consumer electronics, particularly AR/VR, using event-based eye-tracking methods enable lighter, smaller and more comfortable devices. In neuroscience, such systems aid in deciphering visual attention processes and diagnosing neurological disorders, enhancing our understanding of cognitive mechanisms. Our project aims to deploy some popular neural network structures in event-based eye-tracking, compare effectiveness and explore the potential of model interpretability.

\section{Literature Review}

\subsection{Models}
Convolutional Neural Networks (CNNs) have been widely used in event-based eye-tracking research due to their ability to extract features from raw input data. Recurrent Neural Networks (RNNs) are another common architectures in event-based eye-tracking methods. Compared with CNNs, RNNs have the ability to capture temporal dependencies in sequential data by maintaining an internal state or memory. This makes them well-suited for processing event streams. Within event-based eye-tracking, architectures such as Gated Recurrent Units (GRUs), Long Short-Term Memory (LSTM) networks, and Bidirectional LSTMs (Bi-LSTMs) are frequently utilized and demonstrate commendable performance. In our project, we will harness the synergistic potential of CNNs and RNNs by integrating them into a cohesive framework. By combining the strengths of both architectures, we aspire to achieve heightened levels of accuracy [2].

Moreover, one distinguishing aspect of our research is the utilization of data captured from the DVXplorer Mini event camera. This choice sets us apart from many previous studies that rely on different eye-tracking devices or datasets. The DVXplorer Mini event camera offers several advantages, including high temporal resolution, low power consumption, and compact size, making it ideal for capturing fine-grained eye movement data in real-time.

By leveraging data from the DVXplorer Mini event camera, our research benefits from the unique qualities of this device, enabling us to analyze eye movement patterns with unprecedented precision and efficiency. This distinct dataset allows us to explore new avenues in event-based eye-tracking research and provides insights that were previously inaccessible with traditional eye-tracking methods.

In summary, our research stands out by combining CNNs and RNNs in a novel framework while utilizing data from the DVXplorer Mini event camera. This integrated approach, coupled with the unique qualities of the dataset, positions our research at the forefront of interpretable eye-tracking models and opens new possibilities for understanding and analyzing eye movement data.

\subsection{Interpretation}
One main challenge of eye-tracking technique is interpretability. The unclear nature of neural networks poses a significant obstacle to the interpretation and comprehension of models trained on eye movement datasets, hence restricting their usefulness in domains that necessitate explainability [3]. This project aims to develop an optimized, transparent, and interpretable model for predicting pupil coordinates, employing explainable AI techniques to ensure the prediction process is fully understandable. A specific challenge with models trained on eye movement data, as identified in, Challenges in Interpretability of Neural Networks for Eye Movement Data concerns the interpretability of data points. The research has highlighted the difficulties surrounding understanding the relationship between data points and the representations of feature vectors created by the model within its hidden layers. The focus of interpretability tools should, therefore, shift toward dissecting the behavior of these hidden states. By doing so, it becomes possible to examine how these vectors encapsulate the attributes of the input, thereby enhancing the ability to make the model's learning process understandable.  
\section{Methodology}

In recent studies, the amalgamation of Convolutional Neural Networks (CNNs) with recurrent models like Gated Recurrent Units (GRU) or Long Short-Term Memory (LSTM) networks has shown promising results. This fusion offers a unique advantage in feature extraction and temporal modeling, particularly beneficial for tasks involving sequential data analysis. Specifically, in the realm of eye movement detection, CNNs exhibit proficiency in extracting spatial features from event frames, shedding light on intricate eye movement patterns. Concurrently, recurrent models excel in capturing temporal dependencies within sequential data, enabling dynamic learning and adaptation over time.

\subsection{Data Processing and Preparation}

Effective model preparation commences with rigorous data processing to ensure compatibility with the chosen architecture. For this research, our dataset[4] comprises recordings from 13 subjects, spanning multiple recording sessions per subject. These sessions encompass a diverse range of activities, including random movements, saccades, text reading, smooth pursuit, and blinks. Each session generates raw event data, subsequently transformed into event frames for comprehensive analysis. With a dataset volume of approximately 1 GB in compressed .h5 format, diverse challenges emerge in data representation and processing pipelines, contingent upon the chosen methodology and analytical approach.

\begin{figure}[h]
    \centering
    \includegraphics[width=1\textwidth]{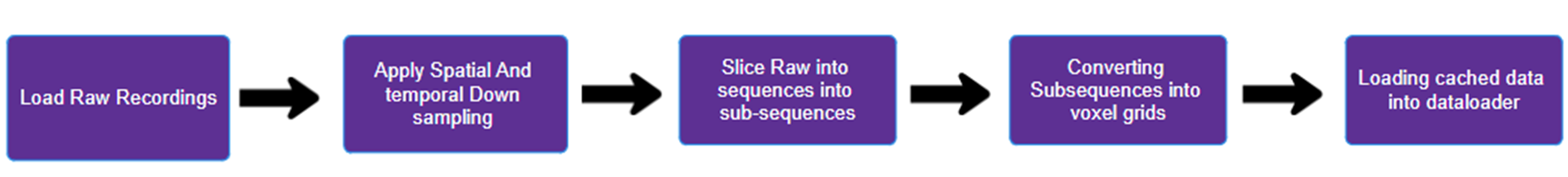} 
    \caption{Figure Showing the flow of data processing steps}
    \label{fig:result-2x1}
\end{figure}

\subsection{Labelling Scheme}

The dataset's ground truth labeling operates at a frequency of 100Hz, encompassing two primary components for each label (x, y, close): pupil center coordinates (x, y) and a binary 'close' indicator denoting eye blinks (0 for eye opening, 1 for eye closing). For the training split, labels are initially provided at 100Hz, offering the flexibility to downsample as per specific requirements. In our provided training sample pipeline, labels undergo downsampling to 20Hz, with the exclusion of the 'close' label by default. Evaluation of the test set is uniformly conducted at a 20Hz frequency, ensuring consistency in evaluation metrics and results. Any modifications to the data loading pipeline should carefully consider this downsampling requirement to maintain evaluation integrity and coherence.

\subsection{Different Approaches we Tried On Our Problem}

\begin{figure}[h]
    \centering
    \includegraphics[width=0.5\textwidth]{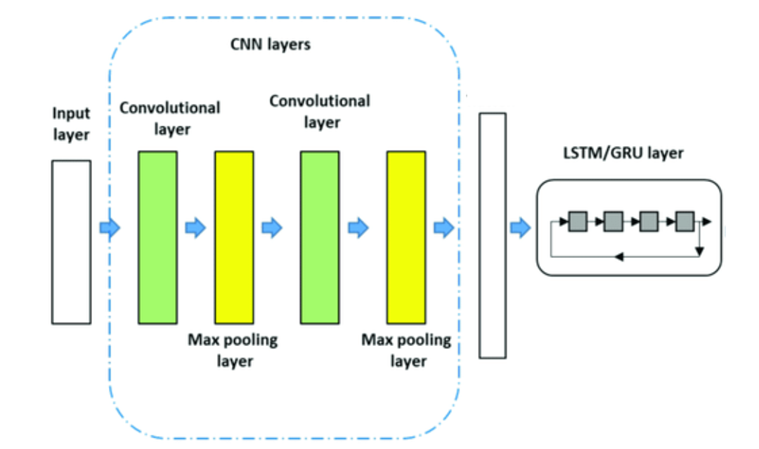} 
    \caption{Figure Showing the high level architecture of our model}
    \label{fig:result-2x1}
\end{figure}

\subsubsection{Model Summary: CNN-GRU for Eye Tracking}
Utilizing a CNN followed by a single-layer GRU, this model architecture integrates spatial and temporal information, making it essential for sequential eye tracking tasks. It is particularly suitable for eye tracking scenarios where understanding both spatial patterns and temporal changes over time is crucial.

\subsubsection{Model Summary: CNN-BiLSTM for Eye Tracking}
The architecture of this model involves employing a CNN with a single-layer bidirectional LSTM (BiLSTM). It offers advantages by combining convolutional and bidirectional recurrent layers, effectively capturing both spatial and temporal features. This makes it well-suited for eye tracking tasks requiring comprehensive analysis of both spatial and temporal dynamics.

\subsubsection{Model Summary: CNN-LSTM for Eye Tracking}
Featuring a CNN followed by a multi-layer LSTM, this model integrates convolutional and recurrent layers to capture spatial and temporal information. It is particularly effective for eye tracking tasks involving intricate temporal dynamics, making it suitable for scenarios where a deep understanding of temporal dependencies is essential.

In the context of eye tracking, each model leverages a combination of convolutional and recurrent layers to address the unique challenges posed by sequential eye movement data. By integrating spatial and temporal information, these models excel in capturing the complex patterns and dynamics inherent in eye movements. This makes them well-suited for tasks requiring precise prediction of pupil center coordinates over time, such as analyzing gaze behavior during visual tasks or monitoring eye movements in real-time applications like augmented reality interfaces.

\section{Evaluation and Results}

The loss function utilized in this study is a weighted mean squared error (MSE), which is defined as:

\[
\text{MSE} = \frac{1}{N} \sum_{i=1}^{N} w_i \cdot (y_i - \hat{y}_i)^2
\]

where $N$ is the number of samples, $y_i$ is the ground truth value, $\hat{y}_i$ is the predicted value, and $w_i$ is the weight assigned to each component. This loss function computes the squared difference between predictions and ground truth, with customizable weights for each component. It is particularly useful when different components of the prediction have varying levels of importance.

For model evaluation, two metrics are used:

\subsection{Pixel Accuracy (p\_acc)}

Pixel accuracy assesses prediction accuracy by comparing predicted pixel coordinates to ground truth within specified tolerance ranges. It is computed as the percentage of correctly predicted pixels within the tolerance range. The formula for pixel accuracy is:

\[
p\_acc = \frac{\text{Number of correctly predicted pixels}}{\text{Total number of pixels}} \times 100\%
\]

where the tolerance range determines how closely the predicted pixel coordinates need to match the ground truth to be considered correct.

\subsection{Pixel Euclidean Distance (px\_euclidean\_dist)}

Pixel Euclidean distance measures the total pixel-wise Euclidean distance between predicted and actual coordinates, indicating how closely the model's predictions match the real eye movement trajectories. It is computed as:

\[
px\_euclidean\_dist = \sum_{i=1}^{N} \sqrt{(y_i - \hat{y}_i)^2}
\]

where $N$ is the number of samples, $y_i$ is the ground truth pixel coordinate, and $\hat{y}_i$ is the predicted pixel coordinate.

These evaluation metrics provide insights into the performance of the eye tracking model, allowing for a comprehensive assessment of its accuracy and effectiveness in predicting eye movement trajectories.

\subsection{Model Performance Comparison}

\begin{table}[h]
\centering
\begin{tabular}{|c|c|}
\hline
\textbf{Model} & \textbf{Accuracy (\%)} \\ \hline
CNN-GRU & 72 \\ \hline
CNN-BiLSTM & 77 \\ \hline
CNN-LSTM & 81 \\ \hline
\end{tabular}
\caption{Model Performance on Test Data}
\label{tab:model_performance}
\end{table}

As shown in Table \ref{tab:model_performance}, CNN-LSTM outperforms CNN-GRU and CNN-BiLSTM with an accuracy of 81\%. This indicates that the CNN-LSTM model provides the most accurate predictions for eye movement trajectories compared to the other models.

\subsection{Explanation of Best Performing Model}

The CNN-LSTM model combines convolutional and recurrent layers, allowing it to capture both spatial and temporal information effectively. This integration enables the model to understand the spatial patterns of eye movements while also capturing the temporal dependencies in sequential eye tracking data.

\subsection{Model Parameters and Techniques}

The CNN-LSTM model was trained using a learning rate of 0.001. Additionally, techniques such as batch normalization and dropout were employed to improve accuracy. Batch normalization helps stabilize the training process by normalizing the input to each layer, while dropout prevents overfitting by randomly dropping neurons during training, thereby promoting generalization.

Important parameters used in training:
\begin{itemize}
    \item Learning rate: 0.001
    \item Batch size: 20
    \item Spatial factor: 0.125
    \item Temporal subsample factor: 0.2
    \item Number of epochs: 200
    \item Pixel tolerances: [5, 10, 15]
\end{itemize}

\section{Additional Work}

\subsection{Overview of Layerwise Relevance Propagation}

Deep neural networks are often labeled as \textit{black boxes }due to the challenging nature of explaining how they generalize. The complexity in interpreting these models becomes particularly obvious with image data, as it involves non-linear processes that transform pixels into abstract feature representations before arriving at a final decision. The ability to interpret and understand a model's reasoning is crucial, enabling subject matter experts to assess and validate a model's performance. 

Within the domain of Explainable AI (xAI), Layerwise Relevance Propagation (LRP), introduced by Bach et al. (2015) emerged as a method aimed at clarifying the inner mechanisms of neural networks. It is characterized as a pixelwise decomposition technique, where a crucial exploratory element, known as a relevance score, is back-propagated through the network. This process aims to pinpoint the impact of an individual pixel in an image x on the prediction outcome f(x) [5]. This method could improve the interpretability of the CNN-LSTM model proposed above. For instance, consider an eye-tracking scenario where the model predicts a user's gaze shifting towards an object on a screen. LRP can be applied post-prediction to highlight the specific pixels or regions in the event frame that had the most significant influence on the model's output. This relevance map could then be visualized as a heatmap overlaid on the original input image, directly showing which areas of the image were significant in determining the gaze direction. 

For LRP to be implemented successfully, it requires access to the models’ parameters, activations, and architecture. In a neural network, the model’s parameters refer to the weights and biases, the activations are the outputs of the activation functions, and the architecture refers to how neurons are connected. After the model is trained, a backward pass of LRP uses the abovementioned parameters to compute each neuron's relevance in a layer. This relevance score is a function of the upper layers' relevances such that no relevance is lost between layers [6]. The general rule given to implement LRP representing the relevance is \(R_j = \Sigma \frac{z_{jk}}{\Sigma_j z_{jk}}\). In this general rule, \(z_{jk}\) represents the contribution of neuron j has towards neuron k to make it relevant [5]. This term is computed in various ways based on the most applicable LRP rules which are discussed below. 

\subsection{LRP Rules}
Various LRP rules exists as there is a continual endeavour to strike a balance between simplicity, stability, and specificity in the interpretation of deep neural networks. This allows subject-matter experts to select the approach that best fits their unique model and application requirements, 

The LRP-0 rule is the most straightforward, distributing relevance linearly based on the proportion of each neuron's contribution to the next layer. It assumes equal contributions of positive and negative values, which can be too simplistic for complex models. 

The next rule, LRP-Epsilon addresses the potential instability in LRP-0 when neurons' contributions are minimal, LRP-Epsilon introduces a term in the denominator. 

Lastly, LRP-Gamma emphasizes positive contributions more heavily than negative ones [6]. The importance of knowing these rules is emphasized below when implementing this method is required. 

\subsection{Challenges with implementing LRP}

\subsubsection{Challenges of a hybrid architecture model}
The main challenge with implementing LRP on a mixture model is the heterogeneity between the layers within the model. In the CNN-LSTM model, the recurrent layers process data in sequences to capture temporal dependencies, and the previous CNN layers apply filters to incoming data to discover spatial features. As a result, each distinct layer requires a particular method to propagate relevances back to the input. While implementations have been completed for both CNN models and RNN models, the combination of both kinds of layers for this eye-tracking application necessitates an implementation from the ground up.

\subsubsection{Challenges of CNN or LSTM models}
It is important to highlight the challenges of applying LRP not only to hybrid CNN-LSTM models but also to standalone CNN and LSTM models individually. This distinction reveals the complex nature of the implementation and how adjustments or combinations of the original rules must be made.

For instance, in an implementation of LRP for a Convolutional Neural network trained on Radar images, a key point was made to alter the implementation to consider weight sharing so that LRP works as expected. The paper notes that LRP needs to account for the shared weights across the convolutional layers. The term \(z_{jk}\) needs to consider the relationship between layers and consequently, capture the dimensions of convolutional kernels and their channels. If this is not done and the typical implementation of LRP is used, the heatmaps produced will not produce easily interpretable results [7]. 

Moreover, there are best practices that should be used to implement LRP for CNNs. For the VGG-16 model which uses convolutional layers, the current best practice is to implement LRP with a “composite strategy”. This means that certain rules are applied to particular convolutional layers. For example, a proposed best-practice approach utilizes the LRP epsilon rule mentioned above to decompose fully connected layers close to the output then the LRP-alpha-beta, and lastly, LRP b-rule [8]. The combination of these rules is significant to consider if the implementation is expected to work well. 

The literature also reveals specific considerations for implementing LRP in LSTM models. The unique configuration of LSTM layers, where data is fed as sequences and neurons interlink in a complex manner, necessitates modifications to the standard LRP-0, LRP-Epsilon, and LRP-Gamma rules. This complexity is accentuated by the topology of the layers where the neurons are connected in two distinct types of connections. They are connected in many-to-one weighted linear connections and two-to-one multiplicative interactions [9]. 

It is established here that there are many considerations to be made and a thorough evaluation of these factors is essential before applying LRP to a deep neural network.

\subsection{Existing Tools for Implementation}
Given the complicated nature of LRP, several libraries have been developed to provide foundational implementations. The links below offer a starting point, although customization according to the specific application is necessary.

\begin{enumerate}
    \item 
    \href{https://github.com/chr5tphr/zennit}{Zennit}: Implemented for use with Pytorch and is still under development. Supports Convolution, Linear, AvgPool, Activation, BatchNorm layers.  

    \item
    \href{https://github.com/albermax/innvestigate}{iNNvestigate}: Implemented for use with Keras and TensorFlow2. 

    \item
    \href{https://github.com/alewarne/Layerwise-Relevance-Propagation-for-LSTMs}{Layerwise Relevance Propagation for LSTMs}: Implemented for use with Tensorflow, Keras and Pytorch. Follows the methodology outlined by Warnecke et al. (2020).

\end{enumerate}

\subsection{Future Considerations}
Advancing this project in the future will involve a deeper investigation into implementing explainable AI techniques, such as LRP, tailored to the CNN-LSTM’s model's complexity. This would include extracting the critical layers and then applying specific LRP rules informed by existing research on LRP within CNNs and LSTMs frameworks. This direction outlines an initial path toward integrating LRP, the current challenges associated with our model necessitate additional exploration for a successful implementation.

\label{others}

\section{Conclusion}

In concluding this research, our examination and comparison of various neural network models have provided insights into the methods of deep learning for eye tracking. Among the architectures evaluated, the CNN-LSTM model achieves the highest accuracy in predicting eye movement. This highlights the effectiveness of combining Convolutional Neural Networks for spatial analysis with Long Short-Term Memory networks for capturing temporal patterns. While our current research focused on exploring model performance, we also considered the future integration of explainable AI (xAI), particularly the exploration of Layerwise Relevance Propagation (LRP) as an area for investigation. To summarize, this report has not only pinpointed the CNN-LSTM architecture as a highly accurate model for eye tracking but also set the stage for further exploration in combining deep learning with xAI.





\newpage
\bibliography{iclr2021_conference}
\bibliographystyle{iclr2021_conference}
[1] S. Stuart, Eye Tracking: Background, Methods, and Applications, vol. 183. New York, NY: Springer US, 2022. doi: 10.1007/978-1-0716-2391-6.\\

[2] Ahmed, Z. A. T., Albalawi, E., Aldhyani, T. H. H., Jadhav, M. E., Janrao, P., \& Obeidat, M. R. M. (2023). Applying Eye Tracking with Deep Learning Techniques for Early-Stage Detection of Autism Spectrum Disorders. Data (Basel), 8(11), 168-. https://doi.org/10.3390/data8110168 
 \\
 
[3] Ayush Kumar, Prantik Howlader, Rafael Garcia, Daniel Weiskopf, and Klaus Mueller. 2020. Challenges in Interpretability of Neural Networks for Eye Movement Data. https://doi.org/10.1145/3379156.3391361  \\

[4] ChrisJudy, Nanashi, Zuowen Wang. (2024). Event-based Eye Tracking - AIS2024 CVPR Workshop. Kaggle. https://kaggle.com/competitions/event-based-eye-tracking-ais2024 \\

[5] Bach S, Binder A, Montavon G, Klauschen F, Müller KR, et al. (2015) On Pixel-Wise Explanations for Non-Linear Classifier Decisions by Layer-Wise Relevance Propagation. PLOS ONE 10(7): e0130140. https://doi.org/10.1371/journal.pone.0130140 \\

[6] Holzinger, A., Saranti, A., Molnar, C., Biecek, P., Samek, W. (2022). Explainable AI Methods - A Brief Overview. In: Holzinger, A., Goebel, R., Fong, R., Moon, T., Müller, KR., Samek, W. (eds) xxAI - Beyond Explainable AI. xxAI 2020. Lecture Notes in Computer Science(), vol 13200. Springer, Cham. https://doi.org/10.1007/978-3-031-04083-2\_2 \\ 

[7] Zang, B., Ding, L., Feng, Z., Zhu, M., Lei, T., Xing, M., \& Zhou, X. (2021). CNN-LRP: Understanding Convolutional Neural Networks Performance for Target Recognition in SAR Images. Sensors (Basel, Switzerland), 21(13), 4536. https://doi.org/10.3390/s21134536 \\ 

[8] Kohlbrenner, M., Bauer, A., Nakajima, S., Binder, A., Samek, W., \& Lapuschkin, S. (2020). Towards Best Practice in Explaining Neural Network Decisions with LRP. CoRR, 1910.09840. https://doi.org/10.1109/ijcnn48605.2020.9206975 \\ 

[9] Arras, L., Montavon, G., Müller, K., \& Samek, W. (2017). Explaining Recurrent Neural Network Predictions in Sentiment Analysis. CoRR, 1706.07206. https://doi.org/10.18653/v1/w17-5221 

[10] Warnecke, A.,  Arp, D.,  Wressnegger, C., \& Rieck, K. (2020). Evaluating Explanation Methods for Deep Learning in Security. 158-174. 10.1109/EuroSP48549.2020.00018. 

\end{document}